%% file: kdd25.tex
\documentclass[sigconf,nonacm]{acmart}
\AtBeginDocument{%
  \providecommand\BibTeX{{%
    \normalfont B\kern-0.5em{\scshape i\kern-0.25em b}\kern-0.8em\TeX}}}

\usepackage{mathpazo} 
\usepackage[utf8x]{inputenc}
\usepackage{rotating}
\usepackage{sankey} 
\usepackage[english]{babel}
\usepackage{tikz} 
\usetikzlibrary{positioning}
\usepackage{pgfgantt} 
\usepackage{float}
\usepackage{graphicx}
\usepackage{datetime}
\usepackage{hyperref}
\usepackage{bookmark}
\usepackage{svg}
\usepackage{booktabs}
\usepackage{tabularx}
\usepackage{float}
\usepackage{soul}
\usepackage{color}
\usepackage{subcaption}
\usepackage{longtable}
\usepackage{tabularray}
\usepackage{rotating}
\usepackage{makecell}

\usepackage{scalerel}
\newcommand{\doubleckmark}{\scalerel*{\begin{tikzpicture}[x=2ex,y=2ex]
\draw[line width=0.72ex] (136:1.25) -- (0,0) -- (51:2.85);
\draw[xshift=2.5ex,line width=0.72ex] (137:0.55) -- (0,0) -- (52:2.85);
\end{tikzpicture}}{W}}

\newcommand{\ckmark}{\scalerel*{\begin{tikzpicture}[x=2ex,y=2ex]
\draw[line width=0.72ex] (136:1.25) -- (0,0) -- (51:2.85);
\end{tikzpicture}}{W}}

\usepackage{todonotes}

\newcolumntype{Y}{>{\centering\arraybackslash}X}

\copyrightyear{2025}
\acmYear{2025}
\acmConference{}
\acmBooktitle{}
\acmPrice{}
\acmDOI{}
\acmISBN{}

\begin{document}

\author{Angelantonio Castelli}
\affiliation{%
  \institution{Booking.com}
  \city{Amsterdam}
  \country{the Netherlands}
}
\email{antonio.castelli@booking.com}
\authornote{The authors contributed equally to this research, listed in alphabetical order.}

\author{Georgios Christos Chouliaras}
\affiliation{%
  \institution{Booking.com}
  \city{Amsterdam}
  \country{the Netherlands}
}
\email{georgios.chouliaras@booking.com}
\authornotemark[1]

\author{Dmitri Goldenberg}
\affiliation{%
  \institution{Booking.com}
  \city{Tel Aviv}
  \country{Israel}
}
\email{dima.goldenberg@booking.com}
\authornotemark[1]

\renewcommand{\shortauthors}{Castelli, Chouliaras and Goldenberg}


\begin{abstract}

With the rapid integration of Machine Learning (ML) in business applications and processes, it is crucial to ensure the quality, reliability and reproducibility of such systems.
We suggest a methodical approach towards ML system quality assessment and introduce a structured Maturity framework for governance of ML. We emphasize the importance of quality in ML and the need for rigorous assessment, driven by issues in ML governance and gaps in existing frameworks.
Our primary contribution is a comprehensive open-sourced quality assessment method, validated with empirical evidence, accompanied by a systematic maturity framework tailored to ML systems. 
Drawing from applied experience at Booking.com, we discuss challenges and lessons learned during large-scale adoption within organizations. The study presents empirical findings, highlighting quality improvement trends and showcasing business outcomes. The maturity framework for ML systems, aims to become a valuable resource to reshape industry standards and enable a structural approach to improve ML maturity in any organization.

\end{abstract}


\title{Maturity Framework for Enhancing Machine Learning Quality}

\maketitle

\section{Introduction}


Machine Learning (ML) has revolutionized the way organizations operate, embedding new capabilities into various domains \cite{booking2021personalization}. However, ensuring its quality has emerged as a paramount concern. This paper addresses the multifaceted challenges surrounding ML quality assessment, reproducibility and governance, presenting a novel Quality Assessment and Maturity Framework for ML systems.


The motivation for this research stems from the pressing need to establish robust mechanisms for assessing and governing ML systems, arising from business requirements for resilience, efficiency \cite{bernardi2019150, huyen2022designing} and external requirements for fairness and regulation \cite{veale2021demystifying, de2021artificial, dixon2023principled}. The inherent complexity of ML systems, coupled with the absence of comprehensive quality assessment frameworks, has raised serious concerns about their reliability and safety \cite{ml-privacy-meter, data-poisoning}. 


A significant body of literature has explored ML quality assessment and governance. Previous research has examined quality attributes, drawn parallels with software engineering maturity models, and introduced governance frameworks \cite{mat-model-software-product, mat-model-for-analysis,ibm-maturity-framework}. However, existing models often lack specificity, applicability, or practical validation. 
Our Quality Assessment and Maturity Framework formalizes quality attributes and substantiates each level with practical evidence. This framework offers a structured approach to monitoring, reproducing and improving ML quality across organizations. To illustrate its applied nature, we showcase its implementation at Booking.com, highlighting the measured impact and lessons learned. 
Our primary contributions can be summarized as follows:
\begin{itemize}
 \item  \textbf{Quality assessment framework}: We introduce a comprehensive framework for assessing ML system quality with a systematic evaluation of critical attributes.
 \item  \textbf{Maturity Framework:} We present a structured maturity framework tailored to ML systems, offering a roadmap to elevate organisations' quality standards.
  \item \textbf{Technical Implementation:} We describe the needed technical work in order to scale, automate and enable ML quality assessments and improvements, and provide an open-source code for implementation.
 \item  \textbf{Real-world Application:} We present the rollout process of our framework at Booking.com, demonstrating applied lessons and showcasing company-wide quality and business improvements.
\end{itemize}
In the subsequent sections, we delve deeper into the development and application of our Quality assessment and Maturity framework, showcasing its intricacies and real-world impact.


\section{Related Work}
Defining software quality is a fundamental challenge in software engineering. One of the first solutions dates back to 1978 through the means of a software quality model (SQM) \cite{McCall}, which is a set of characteristics and their relationships. It provides the basis for specifying quality requirements and evaluation \cite{ISO9126}. The first SQMs \cite{Dromey,Boehm,Grady} are called \textit{basic} and they make global assessments of a software product, while the ones developed for specific domains or applications are characterized as \textit{tailored} quality models \cite{Miguel}.
The quality assessment models (QAM) concept was developed for the purpose of measuring software quality. On top of quality attributes, QAMs also include \textit{metrics} which enable the measurement of the attributes. Examples of such metrics are design metrics, complexity metrics, duplicated code, quality improvement and software reliability \cite{models-measurement-QA,SQAM-systematic-mapping,systematic-open-source-SQAM}. 

The extensive usage of ML systems in production introduced software engineering challenges related to their development and maintenance \cite{large-scale-ML, challenges-in-deploying-ML, a-software-engineering-perspective} and accumulating technical debt over their deployment life-cycle \cite{wamlm2023, sculley2015hidden}. This proliferation of challenges has given rise to the paradigm of Machine Learning Operations (MLOps). 
MLOps introduces a set of best practices to operate ML systems in production at scale \cite{MLOps-Overview,symeonidis2022mlops}. However, despite the rise of MLOps practices the field of quality management for ML systems has received less attention. While there are studies mapping traditional quality management activities to ML systems \cite{quality-management} as well as frameworks for ML quality assessment \cite{towards-guidelines-for-assessing,crisp-ML,quality-assurance-challenges}, none of the existing work provides concrete metrics to be measured.

A concept orthogonal to the quality assessment is the definition of quality standards to determine the maturity of an ML system. The notion of maturity for software applications dates back to 1988 when the first maturity framework was introduced \cite{software-process-maturity-framework}, aiming to improve software development processes, leading to later version of the Capability Maturity Model \cite{CMM}.
The usefulness of maturity models is validated by the development of many variants across time \cite{systematic-mapping-maturity-models, mat-model-software-product}. Since ML systems are software products, the same concept has been applied to ML system development as well \cite{AIMatModelReview}. The general structure of these frameworks is composed of increasing levels of maturity. The concept of \textit{maturity} is often defined at \textit{organization} level \cite{ibm-maturity-framework, mat-model-for-analysis} depending on the level of \textit{automation} \cite{Google-Mat-Model, Microsoft-Mat-Model, towardsMLOps-a-framework} of MLOps activities.
They key differences of our framework with existing ones used as baselines are: a) the maturity levels in our framework are defined on a \textit{system} level, since in practice, organizations own several ML systems at different maturity levels \cite{bernardi2019150}; b) we take into account more quality aspects than automation, such as discoverability, scalability or ownership which are not covered by existing frameworks \cite{Rubric} as discussed in \cite{chouliaras2023best}; c) our framework allows for a quick review of ML systems with respect to internal policies and regulations \cite{veale2021demystifying, de2021artificial, dixon2023principled}; d) we provide different quality criteria for the maturity, based on the system’s criticality, to avoid unnecessary overhead for low criticality systems, which is a concern for any organization.


\begin{table*}[!htbp]
\caption{Full description of quality assessment requirements, and their expected maturity levels. The symbol ``-'' means no requirement, the symbol ``\protect\ckmark'' means the minimal requirement, and ``\protect\doubleckmark'' refers to the full requirement.}
\label{tab:full_qa}
\begin{tblr}{
  colspec = {
    p{0.01\linewidth}
    p{0.05\linewidth}
    p{0.12\linewidth}
    p{0.16\linewidth}
    p{0.32\linewidth}
    p{0.01\linewidth}
    p{0.01\linewidth}
    p{0.01\linewidth}
    p{0.01\linewidth}
    p{0.01\linewidth}}
    , 
    cell{1}{1} = {c=2}{l},
  rowhead = 1,
  cells = {font = \fontsize{7pt}{6pt}\selectfont},
  hlines,
}
 \textbf{Sub-Characteristic} & &\textbf{Minimal req. \ckmark} & \textbf{Full req. \doubleckmark} & \textbf{Reasoning} & \textbf{1} & \textbf{2} & \textbf{3} & \textbf{4} & \textbf{5 }\\
 
\SetCell[r=4]{l} \rotatebox{90}{\textbf{Utility}}  & Accuracy & The ML system outperforms a simple baseline & The ML system outperforms a baseline and its input data are validated & Outperforming a baseline is required to justify the effort of building an ML system \cite{huyen2022designing, poran2022one}. Input data should be validated to avoid problematic system versions deployed in production \cite{google-data-validation}. & \ckmark & \ckmark & \doubleckmark & \doubleckmark & \doubleckmark \\ 

& Effectiveness & Effectiveness is verified with an A/B experiment & Long-term effectiveness is verified by repeating the AB test in 6 months & A/B testing is a reliable way to assess a system's effectiveness \cite{Kohavi-rules-of-thumb, kohavi2022b, bernardi2019150, booking2021personalization}.  & - & - & \ckmark & \ckmark & \doubleckmark \\

& Responsiveness & - & Latency and throughput requirements are met & An ML system, will not have business impact if the latency is too high (real time predictions) or the throughput too small \cite{google-latency}. & \doubleckmark & \doubleckmark & \doubleckmark & \doubleckmark & \doubleckmark \\

& Usability & - & System is deployed in a serving system & An ML system can only have value if it can be effectively used by its potential users.   & - & - & \doubleckmark & \doubleckmark & \doubleckmark \\
\SetCell[r=2]{l} \rotatebox{90}{\textbf{Economy}} &  Cost \mbox{Effectiveness} & - & Revenue from the system is greater than its training and inference costs & A system which costs more to train, maintain,
and serve than the impact of these predictions should not be deployed. & - & - & - & - & \doubleckmark \\

& Efficiency & Basic operations are automated & Resources for training and inference are optimized  & Efficient systems should reach their desired objective with the minimum number of utilized resources \cite{efficient-ml-review}.   & - & - & - &  \ckmark & \doubleckmark \\
\SetCell[r=4]{l} \rotatebox{90}{\textbf{Robustness}} & Availability & - & The deployed service meets its SLAs \cite{wieder2011service} & An unavailable system, cannot achieve its business purpose \cite{sre}. & \doubleckmark & \doubleckmark & \doubleckmark & \doubleckmark & \doubleckmark \\

& Resilience & Up to $30\%$ failed ML pipelines per Q & At most $10\%$ failed ML pipelines per quarter & Automated ML pipelines should not fail frequently to ensure that the most updated model is available for predictions \cite{resilient-ml}.  & - & - & \ckmark & \ckmark & \doubleckmark \\

& Adaptability & The system is partially adaptable  & The system is adaptable (e.g. retrained frequently) & The system operates in a changing environment, hence it should adapt to such changes to avoid losing commercial impact \cite{concept-drift-adaptation}. & - & - & \ckmark & \doubleckmark & \doubleckmark \\

& Scalability & - & The system is deployed and can scale the resources depending on the traffic. & Different ML use cases have different needs in terms of traffic. Given that a system can be used in multiple use cases it is essential to handle traffics of different scale. & - & - & - & - & \doubleckmark \\
\SetCell[r=2]{l} \rotatebox{90}{\textbf{Productionability}} & Repeatability & The pipeline of the ML life-cycle is partially automated & The pipeline repeating the ML life-cycle is fully automated  & Automation of an ML pipeline decreases the overhead of manual actions and minimizes the chances for human error \cite{MLOps-Overview}.  & - & - & \ckmark & \doubleckmark & \doubleckmark \\

& Monitoring & ML performance is being monitored & ML performance, feature drift and metrics are monitored  & ML systems can have many points of failure, it is essential to monitor key indicators to identify performance degradation \cite{MLOps-Overview, monitoring-article}. & - & - & \ckmark & \ckmark & \doubleckmark \\

\SetCell[r=4]{l} \rotatebox{90}{\textbf{Modifiability}} & Maintainability & Code is versioned & Code is versioned and Readability full requirement is met  & The ease of maintenance of an ML system affects the downtime, speed of iteration and hence the commercial impact \cite{maintainability}. & - & \ckmark & \ckmark & \ckmark & \doubleckmark \\
& Modularity & The source code is partially modular  & The code is fully modular, split into components of limited functionality  & Highly modular ML systems allow for changes to be performed in one part of the system without the risk to break another
part of it \cite{modularity}. & - & \ckmark & \ckmark & \ckmark & \doubleckmark \\
& Testability & Test coverage is at least $20\%$  & Test coverage is at least $80\%$ & System test coverage directly affects its robustness and ease of maintenance \cite{software-testing}. & - & \ckmark & \ckmark & \ckmark & \doubleckmark \\
& Operability & The system is deployed on a service  & The system can be disabled, updated and reverted & Production systems might need to have their state altered in case of deployment issues or identified bugs.  & \ckmark & \ckmark & \doubleckmark & \doubleckmark & \doubleckmark \\

\SetCell[r=4]{l} \rotatebox{90}{\textbf{Comprehensibility}} & Discoverability & - & The system is deployed in an accessible registry  & The ability to discover and audit an ML system is essential for ensuring transparency and allowing new users to exploit its value. & - & - & \doubleckmark & \doubleckmark & \doubleckmark \\
& Readability & Meaningful variables names &The code is fully modular, there is a unified code style.  & Easily readable code enhances a system's ability to maintained, modified and extended \cite{code-readablity}. & - & - & \ckmark & \ckmark & \doubleckmark \\
& Traceability & Metadata is \textit{partially} logged & Metadata and artifacts in the ML life-cycle are \textit{fully} logged  & To reproduce production systems it is important to have visibility on the exact conditions they were created and deployed \cite{MLOps-Overview}. & - & - & \ckmark & \doubleckmark & \doubleckmark \\
& Understand- ability & The system has partial documentation & The system has complete documentation  & Systems must be understandable to provide trust to potential users, and allow potential contributors to maintain them \cite{ml-documentation}. & \ckmark & \ckmark & \doubleckmark & \doubleckmark & \doubleckmark \\
\end{tblr}
\end{table*}

\begin{table*}
\begin{tblr}[
  caption = \textbf{Full description of quality assessment requirements},
  entry = {Short Caption},
  label = {tab:full_qa},
]{
  colspec = {
    p{0.01\linewidth}
    p{0.06\linewidth}
    p{0.07\linewidth}
    p{0.18\linewidth}
    p{0.36\linewidth}
    p{0.01\linewidth}
    p{0.01\linewidth}
    p{0.01\linewidth}
    p{0.01\linewidth}
    p{0.01\linewidth}}
    , 
    cell{1}{1} = {c=2}{l},
  rowhead = 1,
  cells = {font = \fontsize{7pt}{7pt}\selectfont},
  hlines,
}
 \textbf{Sub-Characteristic} & &\textbf{Min req. \ckmark} & \textbf{Full req. \doubleckmark} & \textbf{Reasoning} & \textbf{1} & \textbf{2} & \textbf{3} & \textbf{4} & \textbf{5} \\
 
\SetCell[r=5]{l} \rotatebox{90}{\textbf{Responsibility}} & Explainability & - & The system's predictions are explainable  & Explain the mechanism with which the system outputs its predictions is key for gaining stakeholders' trust \cite{explainable-ml-princinples}. & - & - & - & \doubleckmark & \doubleckmark \\
& Fairness & - & The system has been checked against undesired biases and none were identified & ML systems' predictions can be used to take irreversible decisions on behalf of customers, hence it is important that their performance is not affected by undesired biases \cite{ml-fairness}.& \doubleckmark & \doubleckmark & \doubleckmark & \doubleckmark & \doubleckmark \\
& Ownership & - & A team is appointed for maintaining the ML system & Ownership ensures that there is always an appointed individual to maintain the system in case of issues \cite{microsoft-ownership}. & \doubleckmark & \doubleckmark & \doubleckmark & \doubleckmark & \doubleckmark \\
& Standards \mbox{Compliance} & - &Compliance standards, such as PII data handling, are met & Adherence to applied regulatory standards is essential for the long-term viability of an ML system \cite{ml-privacy-meter}.  & \doubleckmark & \doubleckmark & \doubleckmark & \doubleckmark & \doubleckmark \\
& Vulnerability & - & Bots are filtered out from the input data & Existence of bots in data adds noise to the system which harms its performance and pose security risks \cite{data-poisoning}. & \doubleckmark & \doubleckmark & \doubleckmark & \doubleckmark & \doubleckmark \\
\end{tblr}
\end{table*}

\section{Quality of ML systems}
\label{sec:ml_quality}
To define quality, we use a refined version of the quality model tailored for ML systems, introduced in \cite{chouliaras2023best}.
The framework consists of seven quality \textit{characteristics} divided into several attributes, called \textit{sub-characteristics}. Note that we refer to \textit{ML systems} to encompass all the components surrounding the ML \textit{model} artifact, including but not limited to training and deployment code, retraining pipelines, monitoring infrastructure \cite{sculley2015hidden}. The quality model was constructed by shortlisting the quality attributes most relevant for ML systems \cite{List_of_system_quality_attributes}. The completeness of the quality model was verified using published sets of ML practices \cite{sculley2015hidden,Rubric,Amershi,Serban,se_ml_website} and iterative feedback from experienced ML practitioners.

\subsection{Quality Characteristics}
Each of the quality characteristics is described below.

\subsubsection{Utility}
The degree to which an ML system provides functions that meet the needs when used under specified conditions. Utility is important since an ML system has to be useful, in order to bring value to the end users 
\cite{huyen2022designing, bernardi2019150}.

\subsubsection{Economy}
The level of performance relative to the amount of resources used under stated conditions. When describing the quality of an ML system it is essential to consider its cost to impact ratio to ensure positive return on investment \cite{efficient-ml-review}.

\subsubsection{Robustness}
The degradation level suffered by the ML system when exposed to dynamic or adverse events. An ML system should require minimal maintenance in daily operations or in face of small environmental changes  \cite{wieder2011service,resilient-ml, concept-drift-adaptation}.

\subsubsection{Modifiability}
The effectiveness and efficiency with which an ML system can be adapted to changes in environment and requirements \cite{maintainability,modularity,software-testing}.

\subsubsection{Productionizability}
The ease of performing the actions required for an ML system to run successfully in production. An ML system can have value only if it is being used in production for a certain use case \cite{MLOps-Overview, monitoring-article}.

\subsubsection{Comprehensibility}
The degree to which users and contributors understand the relevant aspects of an ML system. Comprehensibility is about building ML systems that are easy to be understood, used, tested, and reasoned about \cite{code-readablity, ml-documentation}.

\subsubsection{Responsibility}
The level of trustworthiness of an ML system. Lack of trustworthiness, in terms of ownership, security, fairness and transparency might pose significant risks for the business \cite{explainable-ml-princinples, microsoft-ownership, ml-fairness, ml-privacy-meter}.

\subsection{Quality assessment}
\label{sec:qa}
To quantify the fulfillment of a quality attribute we assign a \textit{requirement} to each sub-characteristic.
Each attribute may have a \textit{minimal} and a \textit{full} requirement. This is done to allow for incremental quality improvements, and granular quality standards (see \autoref{sec:mat}). Essential sub-characteristics are associated with only full requirements. The requirements along with the reasoning for creating them is provided in \autoref{tab:full_qa}. 
If the full requirements are satisfied, the ML system has \textit{no gap} in the sub-characteristic. Conversely, the gap is \textit{small} when the minimal requirement is fulfilled and is \textit{large} if no requirement is fulfilled. As an example, an ML system that outperforms a simple baseline, but does not have its input data validated, will have a small gap in Accuracy.
By assigning the following numerical value to the three gap types: \textit{no gap}: 0 /  \textit{small gap}: 1 /
\textit{large gap}: 2.
we can introduce the quality score $Q$ as:
\begin{equation}
\label{eq:quality_score}
Q = \bigg \lfloor 100\bigg(1 - \dfrac{\text{\Large $\Sigma$}_s^N g_s}{g_{large}N }\bigg) \bigg \rfloor
\end{equation}
where $g_s$ is the gap value for a given sub-characteristic $s$, $N$ is the total number of sub-characteristics and $g_{large}$ is the value assigned to the large gap. The quality score ranging between 0 and 100, can be used to measure ML system quality, to track improvements and to compare different systems.

The list of characteristics was constructed through literature review and adapted from \cite{List_of_system_quality_attributes} and ISO standard \cite{ISO9126} to the context of ML systems. Due to the subjective nature of such quality models, we validated their usefulness with 40 experienced ML practitioners with qualitative interviewing methods. Yet, the quality sub-characteristics are atomic requirements that in many cases can be automated, such as: \textit{testability} (test coverage percentage), \textit{ownership}, \textit{adaptability} (existence and frequency of retraining pipelines), \textit{availability} (SLA of the system), \textit{effectiveness} (conclusive AB test). In the code (\autoref{section:app}) we provide a function \texttt{automate\_assessment.py} which reads system metadata from an ML registry, and infers the gaps per sub-characteristic based on the provided system metadata to minimize the subjective factor.

\section{Maturity framework}
\label{sec:mat}
%
%

The quality framework provides an overview of the quality of an ML system. To set \textit{quality standards} we implemented a maturity framework composed of the following elements:
\begin{itemize}
    \item Five \textit{levels} of increasing maturity, determined by the fulfilment of the quality attributes.
    \item The notion of \textit{business criticality} defining the importance of a certain system for the organization.
    \item The \textit{expected maturity level} based on the system's business criticality.
\end{itemize}

\subsection{Maturity levels}

We suggest five maturity levels of increasing complexity, defined by a set of different requirements to be fulfilled (see \autoref{sec:qa}). To determine the requirements per level, we rely on the expected quality standards. These depend on which phase in the ML life-cycle the system is in, starting from a proof of concept, to large scale production systems. The maturity levels requirements for each quality sub-characteristic are provided in \autoref{tab:full_qa}. The exact choice of requirements per maturity level depends on the organization's needs and the supporting infrastructure. For example, small organizations with low AI adoption and no safety-critical applications \cite{safety-critical}, may specify a more relaxed set of requirements than the ones proposed in this work. On the contrary, large organizations with safety-critical applications might choose stricter standards, to minimize potential risks. We believe the proposed maturity levels can work well out of the box for most of the large scale organizations with no safety-critical applications. 

\subsection{Business criticality}

Quality improvements and adherence to standards require time and effort. However, not all production systems are equally important for the business, hence, to differentiate among them, we define three business criticality levels \cite{critical-systems} along with their expected maturity levels:

\begin{itemize}
\item \textbf{1 - Proof of concept}: system under experimentation 
\item \textbf{3 - Production non-critical}: production system with a \textit{moderate} business impact (proven by experiment)
\item \textbf{5 - Production critical}: production system with a \textit{large} business impact (proven by experiment)
\end{itemize}
A production system is considered critical if any of the following conditions are met:
\begin{itemize}
    \item The number of requests is larger than the 66th percentile of the distribution of all production systems.
    \item The number of teams or products depending on the system is larger than four.
    \item The total revenue generated by the system is larger than $1\%$ of the yearly revenue.
    \item The system is of strategic importance.
\end{itemize}
The notion of business criticality allows to navigate the trade off between quality and speed of development.
We should note that while maturity levels 2 and 4 are not assigned as required levels, they are important as they provide a sense of progress on the maturity scale and allow for incremental improvements towards the required levels.

\section{ML Quality Python package}
\label{section:app}
\urlstyle{tt}

At Booking.com there are hundreds of ML systems running in production, generating billions of predictions every hour, covering a variety of applications, such as personalization, fraud detection, translation, content generation and many others \cite{bernardi2019150, booking2021personalization}. In order to assess the quality of all systems we built the \textit{ML Quality} Python package and we published it on Github\footnote{You can access the open source version of the package at the following url: \url{https://github.com/bookingcom/ml_quality_maturity_framework}.}.
The core of the package is a class that creates a standardised report for each system (see appendix \ref{appendix:rep_example} for details). The class can be initialized by reading a \texttt{csv} file with the values of the \it gap \normalfont for each quality sub-characteristic and the reason why a certain value was assigned.
This input is used to produce a standardised \it html \normalfont report that can be shared with the practitioner or embedded into a website.
A reproducible example of the creation of a quality and maturity report is stored in the Jupyter notebook \cite{Kluyver:2016aa} called \href{https://github.com/bookingcom/ml\_quality\_maturity\_framework/blob/main/create\_quality\_report.ipynb}{\texttt{create\_quality\_report.ipynb}} in the package's GitHub repository. Typical reports are provided in the \href{https://github.com/bookingcom/ml_quality_maturity_framework/tree/main/images}{\texttt{images}} folder or in the \href{https://github.com/bookingcom/ml_quality_maturity_framework/blob/main/README.md}{\texttt{README.md}} file.

The report provides information about the system's business criticality, expected and actual maturity levels and quality score. A radar chart is used to visualise the quality score over the seven different characteristics. Quality sub-characteristics are listed with the following color-code based on technical gaps and required maturity:
\begin{itemize}
    \item \textbf{Red:} gaps to be filled to reach the next maturity level
    \item \textbf{Orange:} gaps to be filled to reach higher maturity levels up until the required one
    \item \textbf{Yellow:} gaps to be filled to reach maturity levels above the required one
    \item \textbf{Green:} No gaps identified
\end{itemize}
The report provides also standardised recommendation about how to fill the identified technical gaps. The generation and prioritization of the recommendations is based on our framework described in \cite{chouliaras2023best}.

The Python package allows for automated and versioned data collection. Every time an evaluation is performed a directory structure like: \texttt{<team\_name>/<ml\_system\_name>/<date>} is created. The last folder\footnote{The directory structure is generated automatically with the evaluation, as in \href{https://github.com/bookingcom/ml\_quality\_maturity\_framework/blob/main/create\_quality\_report.ipynb}{\texttt{create\_quality\_report.ipynb}}. An example of the folder structure and content is available in the \href{https://github.com/bookingcom/ml\_quality\_maturity\_framework/tree/main/tests/temp\_quality\_reports_test}{tests/temp\_quality\_reports\_test} folder in the GitHub page} contains the reports, the \texttt{csv} file with the gaps, and a Python pickle file containing the class object to guarantee the full reproducibility of the results, even if the initial input file is lost. 
Since all the evaluations are stored in this hierarchical folder structure and can be reproduced at any point in time it is easy to collect data and monitor quality improvements over time, as shown in \autoref{imp_it}.



\subsection{Automated evaluation with ML Registry} 
\label{sec:automation}

The Python package allows the automation of the quality and maturity assessment of any ML system and versioned data collection. According to the maturity of the organization, two levels of automation can be implemented:

\begin{itemize}
\item \textbf{Semi-automated:} the information about an ML system's quality is gathered through a survey form\footnote{See \href{https://github.com/bookingcom/ml_quality_maturity_framework/blob/main/quality_assessment_form.md}{\texttt{quality\_assessment\_form.md}} in the Github page.}. Based on the input collected, the technical gaps are identified manually based on the quality requirements per sub-characteristic shown in Table \ref{tab:full_qa}. Reports are automatically generated via the Python package.
 
\item \textbf{Fully-automated:} requires a registry to store the ML systems' metadata. The ML Quality Python package allows to fetch the registry data and infer the gaps per model without human intervention\footnote{\textit{Readability} and \textit{modularity}, are the only two quality aspects still requiring a human in the loop to evaluate the ML system code base and define the technical gap value. Full automation can be achieved using tools like \href{https://www.sonarsource.com/products/sonarqube/}{Sonarqube} or LLMs.}. This is implemented in the \href{https://github.com/bookingcom/ml\_quality\_maturity\_framework/blob/main/ml\_quality/assessment\_automation.py}{\texttt{ml\_quality/assessment\_automation.py}} module of the package. 
\end{itemize}

In parallel with the development of the framework, Booking.com introduced the \textit{ML Registry}: a centralised repository where all the ML systems in the company are registered with their most relevant information, e.g. performance over a test dataset, training pipeline, code repository link, documentation, etc.
The development of the Booking.com ML Registry allowed to fully automate the evaluation of all the ML systems deployed in production.
The information from the registry is persisted in a daily snapshot table for analytics purposes. An asynchronous job reads the daily table with monthly cadence and generates the \texttt{csv} files used as input for creating the reports. 
The synergy between the \textit{ML Registry} and the \textit{ML Quality} package has proven to be very powerful to implement governance of ML systems. The registry keeps track of all ML systems metadata and is used to maintain a real-time dashboard with ML systems' quality aggregated over all levels of the organization, i.e. departments, tracks or teams. Department leaders receive a monthly report with the information around quality and maturity of their systems, and how they compare with respect to company average.




\section{Maturity Framework Rollout}

The quality and maturity framework started as a centralised effort at Booking.com to address the quality and compliance of hundreds of ML systems across multiple, independent product teams \cite{bernardi2019150, booking2021personalization}. This required leadership support, alongside strong community effort of individuals across teams.

Our two-year rollout strategy was iterative, including conceptualization, policy implementation, and tooling. We launched ML governance policies emphasizing \textit{ethical use of ML}, \textit{ownership}, \textit{auditability}, and \textit{high-quality ML development}. We provided tools for data and code quality improvements, observability, and an ML registry. Finally, we engaged the ML community through pilot assessments and company-wide alignment.


%
\subsection{Challenges and special cases}

During the rollout of the quality and maturity framework, we faced multiple obstacles related to technical and organizational aspects, alongside with peer feedback on the framework. Some of these key challenges and special cases are:

\subsubsection{ML systems integration and data collection}
One of the first challenges was to identify all the ML systems used in the company. While the vast majority of the ML systems were served via a centralised platform \cite{bernardi2019150}, some relied on different infrastructure and were missing crucial documentation for their assessment and reproducibility. This required a company-wide data-gathering effort, to ensure full coverage of the existing systems via the centralised ML Registry (see \autoref{sec:automation}).

\subsubsection{ML systems granularity}
\label{sec:granular}
While a classical scenario assumes a single ML model for a specific application, in practice, this one-to-one relation was not the norm. A single ML model can be used in different use cases and multiple models can be used in a single application.
That is why we suggest to focus the evaluation on the \textit{ML system}, i.e. the model and its application plus all the necessary engineering steps for its serving.
To ease the review process of similar and continuously developing systems, we introduced the concept of \textit{ML system family}, requiring a single evaluation for a specific set of systems sharing quality attributes with the same level of fulfillment.

\subsubsection{ML systems ownership}
To ensure continuity and resilience, all the ML assets in the company were assigned to an owning team. According to the \textit{ownership model}, the ownership is automatically reallocated in a case the team is dissolved. However, we have faced several types of problems with this approach: 
\begin{itemize}
    \item Teams that were not aware of the systems assigned to them (e.g. automatic inheritance from other teams).
    \item Teams that were not capable of maintaining the system as they didn't have ML craft individuals to support.
    \item Teams that did not have priorities around the maintenance work for the assigned systems.
    \item Ownership was automatically allocated to a senior leadership representative, with no technical context.
\end{itemize}
One of the main reasons for the problem was the historical mapping of the ownership by model artifacts and not by ML systems (as described in \autoref{sec:granular}).
In order to resolve ownership, maintenance and potential cleanups, we escalated the issue to product directors, to re-allocate the ownership and the maintenance work within their organizations.

\subsubsection{Legacy ML systems}
 ML systems lacking a clear owner or used in a legacy, unmaintained application required special treatment. The current business value of such systems was unclear, and the owning teams may not have the incentive to improve their quality.
In such cases, we suggested a ``block-out'' protocol, i.e. to compare the system's business contribution with a simple benchmark through an A/B experiment. It led to a cleanup of dozens of ML systems previously believed to bring value, which resulted in lower technical debt and serving costs.
In fact, some of the experiments led to an improvement in business metrics, suggesting that stale systems could have negative effects if not benchmarked continuously.

\subsubsection{Maturity framework validity and adjustment}

A key obstacle of the maturity framework rollout was its adoption by the ML practitioners community, i.e. the pro-activeness to perform the quality evaluations and implement the recommended improvements.
Besides some general pushback, we received structural feedback on the proposed framework.

Quality requirements, such as unit tests coverage, frequency of retraining and continuous comparison with a baseline, were challenged to be arbitrary and too strict. We substantiated our requirements and thresholds by taking them from the quantiles of the empirical distribution of the ML systems in the company (e.g. for training cost) or from well-known industry standards  (e.g. benchmarking against simple baselines) \cite{huyen2022designing}.
For some ML systems we had to adjust the framework to accommodate special cases, as described in \autoref{sec:unique}.

Another source of criticism was the high expectation of quality attributes for specific maturity levels (see \autoref{tab:full_qa}). Some of the requirements were perceived as unnecessary during a POC phase, or too tedious compared to the overall effort.
We addressed this feedback by educating the teams on the subject. For instance, an internal training on ML unit-testing, reduced the entrance barrier, and provided more evidence to prioritise the quality improvement task. At the same time, we adjusted the framework, narrowing it to the most necessary requirements on each level, as a response to comprehensive peer-review feedbacks on the framework.



\subsubsection{Domain specific adjustments}\label{sec:unique}
While our initial maturity framework was tailored to classical ML systems such as: classification, forecasting or recommendation, it required additional adaptations to specific applications.
Computer Vision and Natural Language Processing models \cite{wang2023mumic}, for instance, have different expectations on continuous retraining and baseline comparison, due to the need of human labeling. Similarly, causal ML systems such as uplift models \cite{albert2022commerce} or bandits applications \cite{booking2021personalization}, required to adjust the benchmarking and monitoring criteria for counterfactual evaluation, to allow what-if comparisons with unobserved data.
The rise of Large Language Models (LLM) based applications \cite{wang2023text2topic}, required to address both individual LLM components (base models) and the system as a whole, allowing to evaluate the application from multiple perspectives. GenAI systems share the same quality attributes with traditional ones, however their scale and complexity require modifications in their assessment criteria. We find the main differences in the following aspects:
\begin{itemize}
    \item High fine-tuning and inference costs for GenAI systems require taking into account all potential cost aspects \cite{zheng2023judging} while evaluating cost effectiveness. 
    \item Given the pre-trained nature of foundation models, the requirement for adaptability is shifting from updating the model’s training data, to updating the model’s evaluation data \cite{systematic_approach_benchmark,jie2022alleviating}.
    \item Monitoring is an open challenge for GenAI systems, since most of the labels needed require human annotation and might not be constantly available \cite{how_to_evaluate_llms}. 
    \item GenAI systems pose new security threats, hence mitigation techniques like canary tokens are required \cite{owasp_top_10}.
\end{itemize}

\subsection{Lessons learned}
We provide an overview of the key lessons learned during the rollout of the maturity framework.

\subsubsection{Community Effort}
\label{sec:community}

Rolling out the maturity framework in a large corporate implies a significant culture change, and requires collaboration at multiple levels of stakeholders. At Booking.com we had previous experience democratising online experimentation, and fostering experimentation culture \cite{kaufman2017democratizing}, therefore we relied on many community-driven learnings from this initiative.
Besides strong leadership support, it is crucial to onboard the large community of the individual practitioners to the mission. This was done by conducting trainings, pro-active role-modeling, recruitment of community ambassadors and recognition.
While rolling out evaluations and providing critical findings may trigger push-backs and defensive behaviour, it is important to avoid ``judging'' but rather focus on the improvement opportunities.
Moreover, in order to scale manual tasks in the quality review process (such as reviews of documentation completeness or code quality), we introduced a peer-review mechanism, allowing the process to scale without the direct involvement of the facilitator.

\subsubsection{Tooling}
Scaling the quality assessments required automated and centralised tooling to register and evaluate all the ML systems as described in \autoref{sec:automation}.
An important part of the process, was not only the evaluation of the ML systems, but also the facilitation of the quality improvements. By leveraging the capabilities of the Booking.com central ML platform \cite{bernardi2019150}, we introduced new tools to support the registration of ML systems, code quality checks, unit testing and version control, ML experimentation management and ML observability  \cite{chorev2022deepchecks}. Enabling the access to these capabilities significantly contributed to improving ML quality at scale. Such an ML platform can focus on generic MLOps capabilities, or provide domain specific suite of evaluation, business-rules adjustments, benchmarking and model comparison, such as in recommendations \cite{ross2022democratizing, albert2022commerce}, or Vision and NLP models \cite{wang2023text2topic}.

\subsubsection{Data driven progress tracking}

An important part of the quality improvement process was the enablement of progress tracking. The creation of an ML governance dashboard, and recurrent reports to teams and higher management, created visibility, high-level mapping of the gaps, and made the improvement progress comparable. The increased visibility for each department, motivated leaders to allocate resources to quality improvements.  Moreover, the reports served as an implicit leaderboard, suggesting additional motivation for the teams to invest in quality improvements.

\subsubsection{Bringing value} \label{sec:bvalue}
In some cases, an extensive investment in technical debt and quality improvements received pushbacks from the teams, as they did not provide an explicit business value proposition. Therefore, we provided multiple examples showcasing the potential business gains from quality improvements. We divided them into three categories:

\begin{itemize}
    \item \textbf{Direct:} contributions to the business metrics
    \item \textbf{Indirect:} enablement for faster future development
    \item \textbf{Retrospective:} guardrails to prevent negative impact caused by low-quality ML systems
\end{itemize}
Specific examples are provided in \autoref{sec:value}. 


\subsection{Value of quality improvements examples}
\label{sec:value}
\subsubsection{Quality review and improvements}
\label{sec:casestudy}

Flights reservation is a new product on Booking.com, and as such it was deployed gradually, while reusing existing components. One of them was the \textit{departure airport recommendation model} \cite{ross2022democratizing,booking2021dataset}. 
The model had a proven positive impact on the recommendations. However, due to ownership gap between three teams (front-end use case owners, model developers and underlying dataset maintainers), when one of the data dependencies failed, the model started to return trivial recommendations (the most popular airport in the dataset). After a review of the model, gaps in ownership, adaptability, testability, monitoring and robustness, were identified. These abilities could raise the issue earlier and reduce negative impact on the product.

\subsubsection{Value of retraining}
\label{sec:retraining}
During our assessments we identified production systems with no scheduled retraining regime. To highlight the drawbacks of model staleness, we conducted an exercise: We ran non-inferiority A/B experiments \cite{d2003non,kornilova2021mining} comparing the production systems with simple, low-cost baselines. In most cases, we found that using a low-cost baseline is non-inferior to the production system, which indicates that the performance of the system has degraded, since the initial tests against baselines were positive. Thus, we created a company-wide policy of periodic benchmarking against a low-cost baseline using superiority A/B experiments and made it an integral part of the quality assessment framework (see Effectiveness requirements in \autoref{tab:full_qa}).

\begin{figure}[b]
    \centering
    \includegraphics[width=0.9\columnwidth]{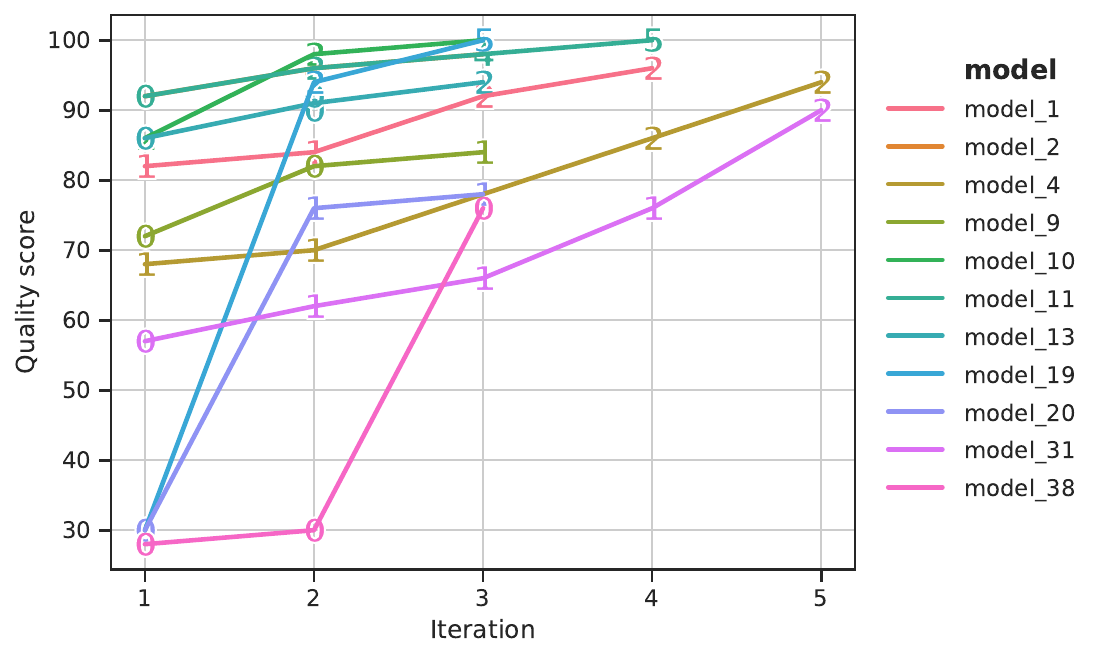}
    \caption{Quality (Y-axis) and maturity (markers) score of a selected subset of systems over iterations.}
    \label{imp_it}
\end{figure}

\subsubsection{Efficiency Improvements}
\label{sec:efficiency_improvements}
We identified a pattern with several seasoned production systems. While their overall quality, and business impact were high, the efficiency of their retraining pipelines was lacking. In particular, the training and hyperparameter optimization steps, had a much longer duration than the 80th percentile of the overall distribution (contrary to the requirement of Efficiency \autoref{tab:full_qa}). The teams worked on improving the efficiency of the pipelines, significantly reducing their duration. As an example, a team owning a business critical system reduced the training time from 7 hours to less than 1 hour without performance loss, by applying efficiency optimizations. This led to savings in computation resources and faster turnaround time for model iteration. For another resource heavy system, the team replaced the hyperparameter tuning regime \cite{parzen-estimator}, reducing the duration from 20 hours to less than 1 hour, without compromising performance.




\section{Outcomes}

\begin{figure}[b]
    \centering\includegraphics[width=0.85\columnwidth]{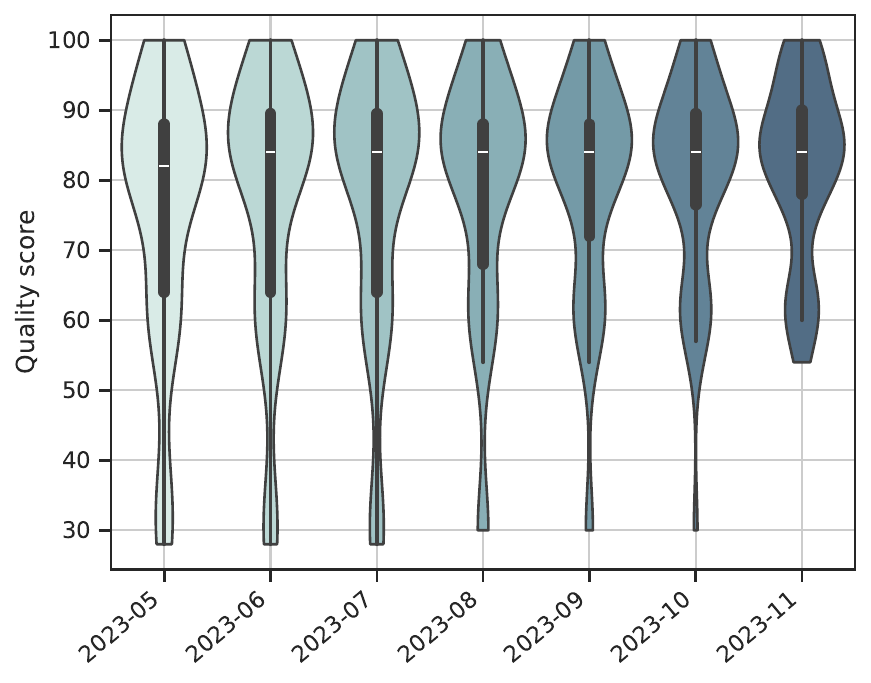}
    \caption{Violin plot of the ML quality score for different months. Each shape represents the score distribution across  evaluated models. The white mark indicates the median.}
    \label{figure:avg_qual}
\end{figure}

The introduction of the maturity framework, and its implementation and company-wide rollout, generated a greater engagement over ML system ownership and reliability in the company. \autoref{imp_it} shows the quality improvement of several ML systems over a number of cycles of assessments and implementation of the recommended best practices. The figure shows the two dimensions of the framework: the quality score (see \autoref{eq:quality_score}) and the maturity level. Certain requirements are essential to move from one maturity level to the other (see \autoref{tab:full_qa}). Therefore, even a significant improvement in the quality score, may result in a low maturity level, as seen in the figure. 
We observed a consistent quality improvement for all the ML systems included in the voluntary rollout. 

\autoref{figure:avg_qual} shows the quality score distribution over time, demonstrating a steady increase of the central value of quality score and a progressive shrinkage of the low-quality tail.

\begin{figure}[t]
    \centering
    \includegraphics[width=0.95\columnwidth]{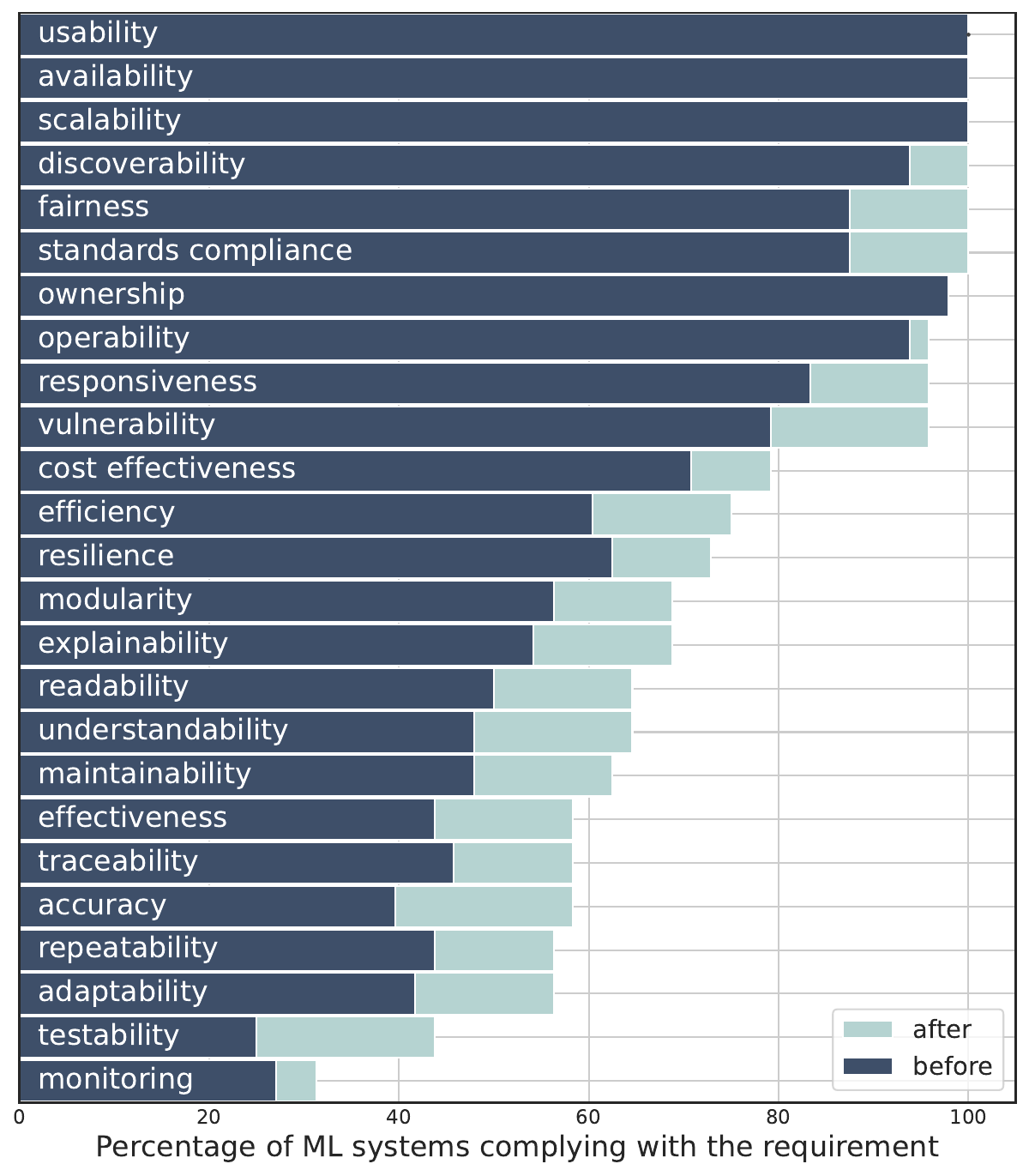}
    \caption{Percentage of systems complying with the requirements  before and after the framework rollout}
    \label{figure:gap_subchar}
\end{figure}

The quality and maturity framework allowed a data-driven prioritisation of the efforts at company level. For each quality sub-characteristic \autoref{figure:gap_subchar} shows the percentage of systems complying with the requirements (without technical gaps) before and after the rollout of the framework.
The centralised serving platform guaranteed that all systems served in production to have no technical gaps in \textit{availability}, \textit{usability} and \textit{scalability}.
We observed a significant improvement over almost all sub-characteristics, and especially for \textit{accuracy}, \textit{testability}, \textit{readability}, \textit{understandability}, \textit{maintainability} among the others. 

We can see how \textit{monitoring} had a major technical gap. Therefore we started a centralised effort, by providing a complete ML Observability platform with key monitoring capabilities. The solution is deployed in production and it will remove the monitoring technical gaps for all ML systems without any extra effort from the teams.
The usefulness of the framework has been validated through the following key findings:
\begin{enumerate}
    \item Higher quality scores were observed in teams with multiple ML practitioners, highlighting the correlation between quality scores and team proficiency.
    \item Identified quality gaps were improved over time, suggesting that they were accepted as beneficial and prioritized by the teams. Moreover, it led to business improvements as described in \ref{sec:value}.
    \item We found legacy ML systems with low quality scores showing constantly degraded business impact and usability, while systems with high scores showing sustained business impact and re-usability. 
\end{enumerate}

\section{Conclusion}

Our framework offers guidance for organizations to address multifaceted challenges of ensuring quality and reproducibility of their ML systems. 
We demonstrated an applied framework implementation at Booking.com, illuminating the potential to drive company-wide quality enhancements, ML reproducibility and substantial business improvements. We showcased the challenges and learnings of the framework rollout process from its conceptualization to becoming a company-wide standard, leading to significant quality improvements across all the aspects, resulting in business value and efficient development of ML in the company. There is still room to improve the ease-of-use, robustness and coverage of the assessments, by introducing more tooling, automation and driving culture change.
We provide tools and a structured pathway for organizations to elevate their standards. With further empirical validation, this framework holds promise in becoming an industry standard, fostering trust and reliability in the ever-evolving landscape of ML systems.
Moreover, with the rise of new families of ML applications, such as Large Language Models, causal and reinforcement learning models, that may be treated as ``black box'' solutions, there is a growing need for guiding principles for quality and governance. Therefore, our suggested framework should be considered as an initial stepping stone to the practice of ML quality, and may be extended with the rapid development of ML systems and adjusted to the specific needs of each organization.




\pagebreak
\balance
\bibliographystyle{ACM-Reference-Format}
\bibliography{bibliography}

\pagebreak

\appendix
\onecolumn

\section{Report examples}
\label{appendix:rep_example}
In this appendix we show examples of a full report generated with the \textit{ML quality} Python package
\input{includes/report_examples}




\end{document}

%% file: includes/report_examples.tex
\begin{figure*}[h]
 \begin{subfigure}{0.48\textwidth}
    \centering
    \includegraphics[width=\linewidth]{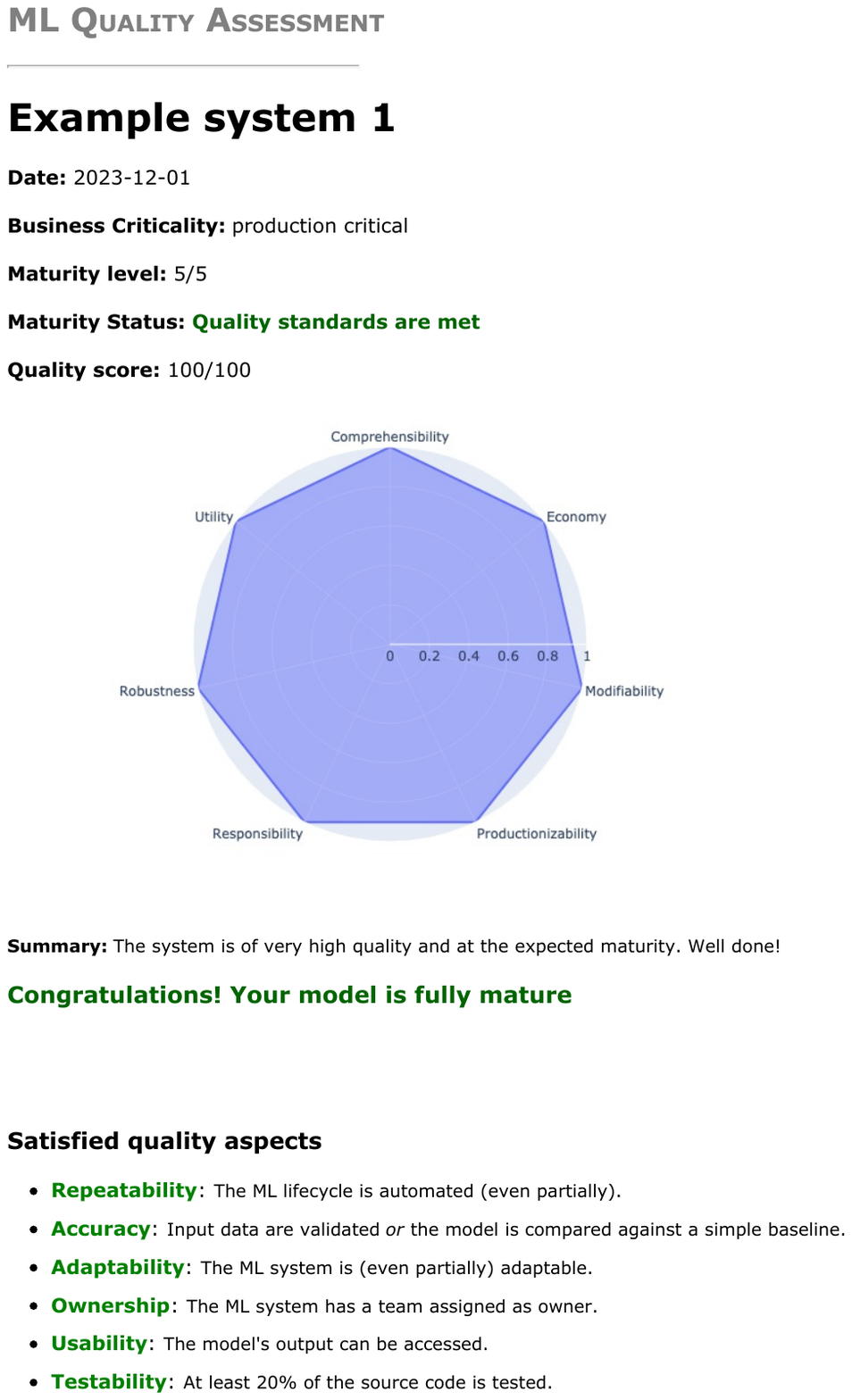}
    \caption{Report page 1}
\end{subfigure}
  \hspace*{\fill}   
\begin{subfigure}{0.48\textwidth}
    \centering
    \includegraphics[width=\linewidth]{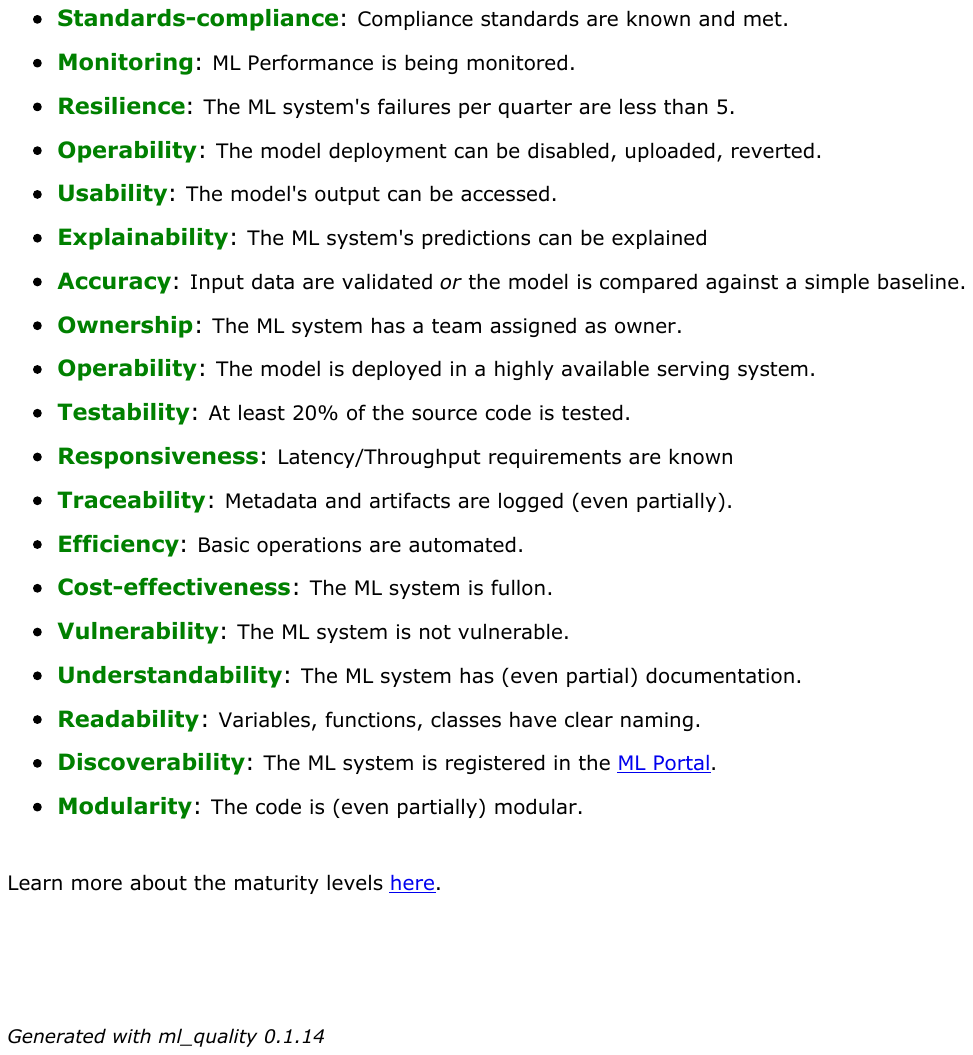}
    \caption{Report page 2}
\end{subfigure}
\caption{Example of a report for a fully mature system. No technical gaps are present, all the fulfilled quality attributes are listed in green}
\label{report_100}
\end{figure*}

\begin{figure*}[h]
 \begin{subfigure}{0.48\textwidth}
    \centering
    \includegraphics[width=\linewidth]{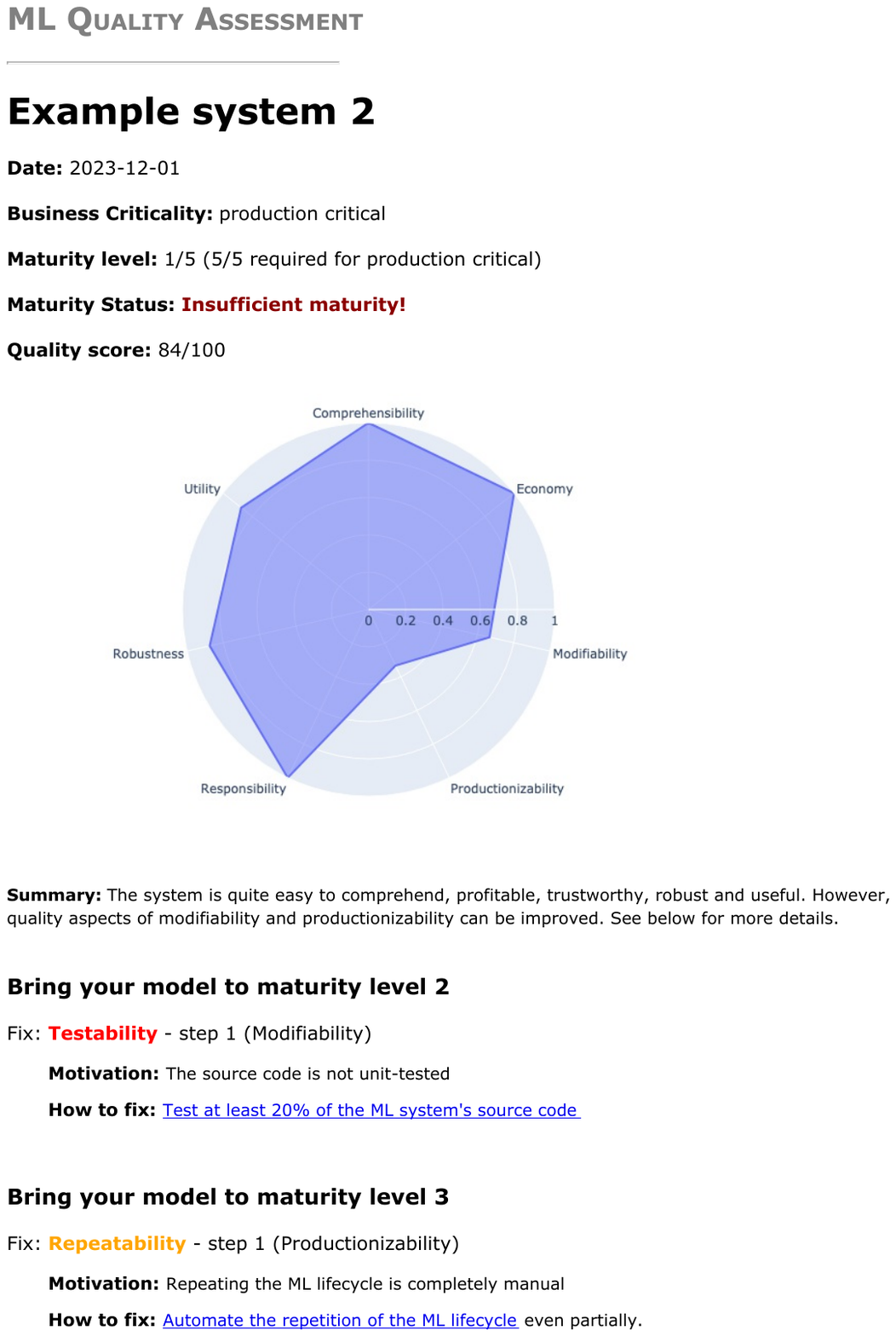}
    \caption{Report page 1}
\end{subfigure}
  \hspace*{\fill}   
\begin{subfigure}{0.48\textwidth}
    \centering
    \includegraphics[width=\linewidth]{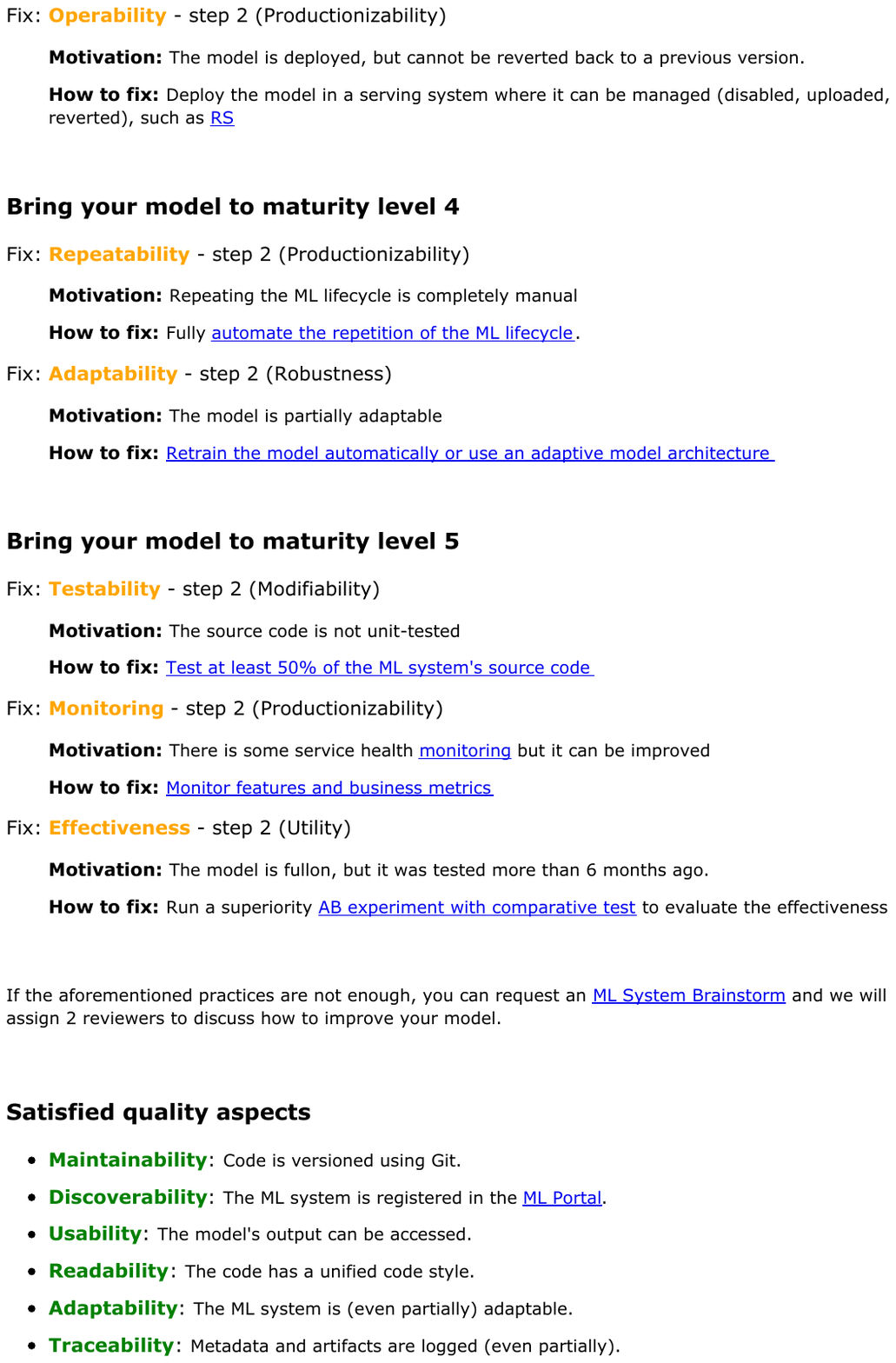}
    \caption{Report page 2}
\end{subfigure}
\caption{Example of a report for a system of maturity level 1. The gaps to be fulfilled to pass to the next maturity level are shown in red. The quality attributes to be fulfilled for the subsequent maturity levels are shown in orange. Below each quality attribute the user can see both the motivation of a certain technical gap and a recommendation to remove it.}
\label{report_84}
\end{figure*}